\def\year{2021}\relax
\documentclass[letterpaper]{article} % DO NOT CHANGE THIS
\usepackage{aaai21}  % DO NOT CHANGE THIS
\usepackage{times}  % DO NOT CHANGE THIS
\usepackage{helvet} % DO NOT CHANGE THIS
\usepackage{courier}  % DO NOT CHANGE THIS
\usepackage[hyphens]{url}  % DO NOT CHANGE THIS
\usepackage{graphicx} % DO NOT CHANGE THIS
\usepackage{graphicx}  % DO NOT CHANGE THIS
\usepackage{natbib}  % DO NOT CHANGE THIS AND DO NOT ADD ANY OPTIONS TO IT
\usepackage{caption} % DO NOT CHANGE THIS AND DO NOT ADD ANY OPTIONS TO IT
\usepackage[switch]{lineno}  

\usepackage{amsfonts}
\usepackage{amsmath} 
\usepackage{subcaption}
\usepackage{color}
\usepackage{algorithm}
\usepackage{algpseudocode}
\usepackage{tcolorbox}
\usepackage{wrapfig}

\renewcommand{\vec}[1]{\boldsymbol{#1}}
\newcommand{\x}[0]{\vec x}
\newcommand{\y}[0]{\vec y}

\title{Supervised Training of Dense Object Nets using Optimal Descriptors for Industrial Robotic Applications}

\author{
  Andras Gabor Kupcsik\textsuperscript{\rm 1}, Markus Spies\textsuperscript{\rm 1}, Alexander Klein\textsuperscript{\rm 2}, Marco Todescato\textsuperscript{\rm 1}, \\ Nicolai Waniek\textsuperscript{\rm 1}, Philipp Schillinger\textsuperscript{\rm 1},  Mathias B\"{u}rger\textsuperscript{\rm 1}
  \\
}
\affiliations{
	 \textsuperscript{\rm 1}Bosch Center For Artificial Intelligence\\
	 Renningen, Germany \\
	 firstname(s).lastname@de.bosch.com\\
	 \textsuperscript{\rm 2}Technische Universit\"{a}t Darmstadt\\
	 Darmstadt, Germany \\
	 alex.klein@stud.tu-darmstadt.de
}

\begin{document}

\maketitle

%===============================================================================

\begin{abstract}
    Dense Object Nets (DONs) by \citet{Florence2018} introduced dense object descriptors as a novel visual object representation for the robotics community. 
    It is suitable for many applications including object grasping, policy learning, etc. 
	DONs map an RGB image depicting an object into a descriptor space image, which implicitly encodes key features of an object invariant to the relative camera pose.
	Impressively, the self-supervised training of DONs can be applied to arbitrary objects and can be evaluated and deployed within hours. 
    However, the training approach relies on accurate depth images and faces challenges with small, reflective objects, typical for industrial settings, when using consumer grade depth cameras.
    In this paper we show that given a 3D model of an object, we can generate its descriptor space image, which allows for supervised training of DONs. 
    We rely on Laplacian Eigenmaps (LE) to embed the 3D model of an object into an optimally generated space. 
    While our approach uses more domain knowledge, it can be efficiently applied even for smaller and reflective objects, as it does not rely on depth information. 
    We compare the training methods on generating 6D grasps for industrial objects and show that our novel supervised training approach improves the pick-and-place performance in industry-relevant tasks.
    
\end{abstract}

%===============================================================================

\begin{figure*}[th]
	\centering
	\begin{minipage}[b]{0.2\linewidth}
		\includegraphics[width=1\textwidth]{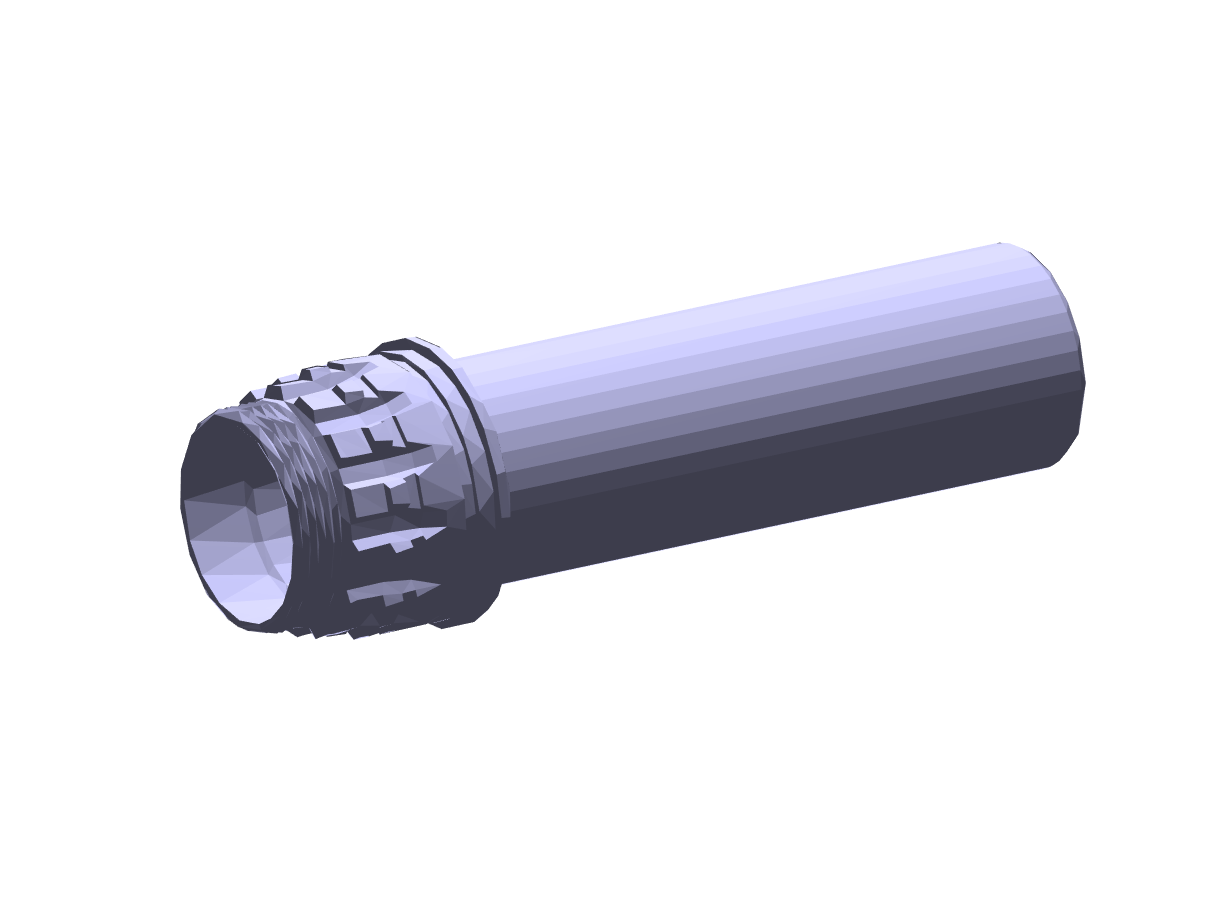} \\
		\vspace*{-1.2cm}
		\includegraphics[width=1\textwidth]{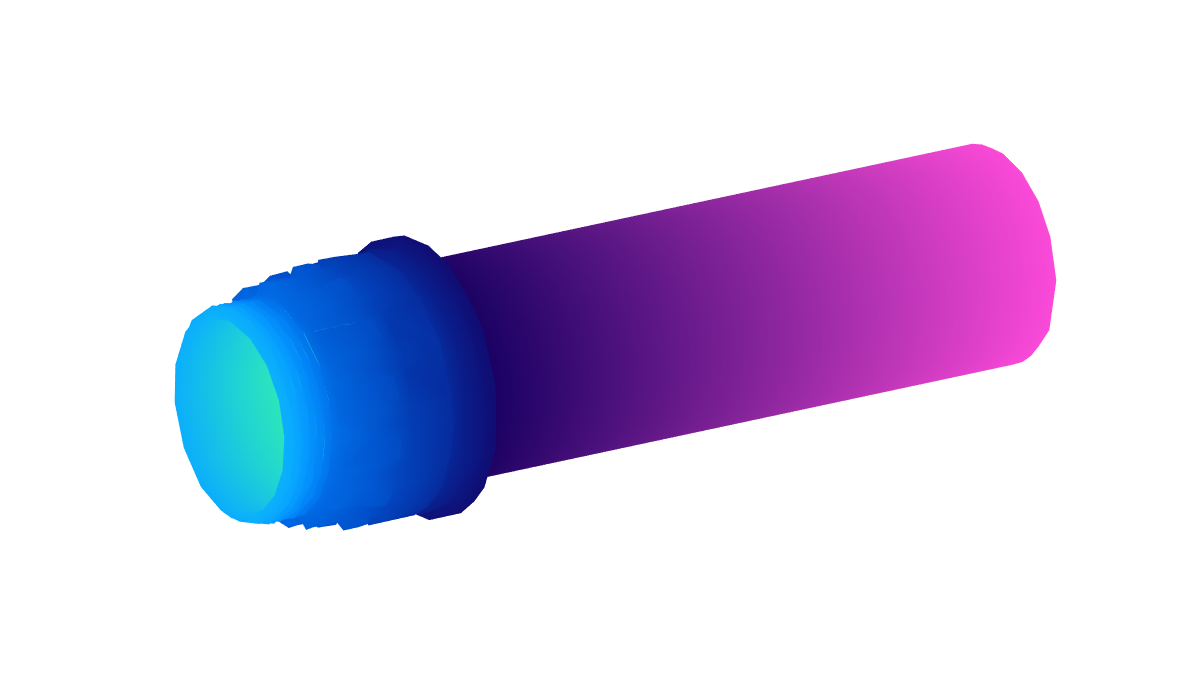}
	\end{minipage}
	\includegraphics[width=0.3\textwidth]{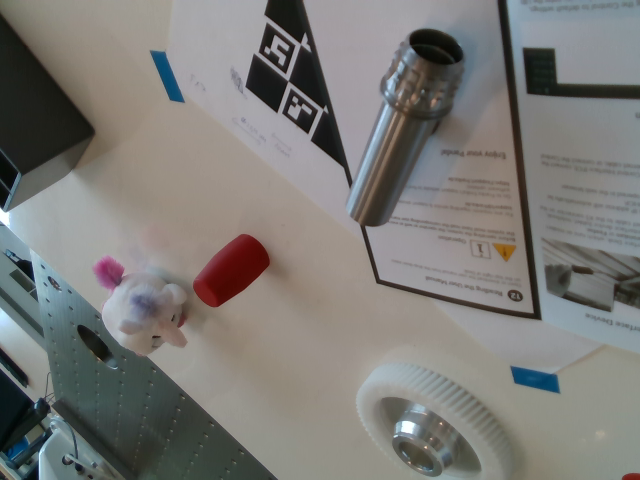}
	\includegraphics[width=0.3\textwidth]{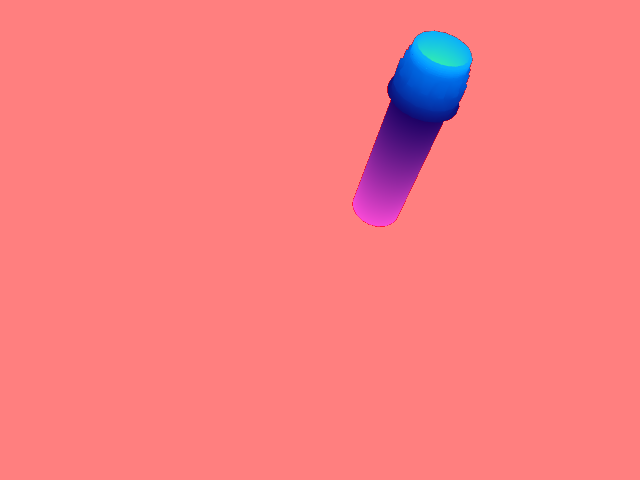}
	\caption{Illustration of the proposed approach (best viewed in color). Given the 3D model of an object \textbf{(left top)} we generate its optimal descriptor space embedding using Laplacian Eigenmaps \textbf{(left bottom)}. Then, for a given input image \textbf{(middle)} with known object pose we render its descriptor space target \textbf{(right)}. For illustration purposes we use $3$ dimensional descriptor space.} 
	\label{fig:illustration}
\end{figure*}
\section{Introduction}
\label{sec:introduction}

Dense object descriptors for perceptual object representation received 
considerable attention in the robot learning community 
\cite{Florence2018, Florence2020, Sundaresan2020}.
To learn  and generate dense visual representation of objects, Dense Object Nets 
(DONs) were proposed by \citet{Florence2018}. DONs map an $h \times w \times 3$ 
RGB image to its descriptor space map of size $h \times w \times D$, where 
$D\in \mathbb{N}^+$ is an arbitrarily chosen dimensionality. 
DONs can be trained in a self-supervised manner using a robot and a wrist-mounted consumer grade RGBD camera, and can be deployed within hours. 
Recently, several impactful applications and extensions of the original 
approach have been shown, including rope manipulation \citep{Sundaresan2020}, 
 behavior cloning \citep{Florence2020} and controller learning \citep{Manuelli2020}.

DONs can be readily applied to learn arbitrary objects with relative ease, including non-rigid objects.
The self-supervised training objective of DONs uses contrastive loss \cite{Hadsell2006} between pixels of image pairs depicting the object and its environment. 
The pixel-wise contrastive loss minimizes descriptor space distance between corresponding pixels (pixels depicting the same point on the object surface in an image pair) and pushes away non-correspondences. 
In essence, minimizing the contrastive loss defined on descriptors of pixels leads to a view invariant map of the object surface in descriptor space. 

The contrastive loss formulation belongs to the broader class of projective nonlinear dimensionality reduction techniques \cite{VanDerMaaten2009}. 
The motivation behind most of these methods is to perform a mapping from an input manifold to a typically lower dimensional output space while ensuring that similar inputs map to similar outputs.
While the contrastive loss formulation relies on a similarity indicator (e.g., matching vs. non-matching pixels), other techniques also exploit the \emph{magnitude} of similarity implying local information about the data (e.g., ISOMAP \cite{Tenenbaum2000} and LE \cite{Belkin2003}).

 While generating similarity indicators, or pixel correspondences, is suitable for self-supervised training, its accuracy inherently depends on the quality of the recorded data.
Based on our observations, noisy depth data can deteriorate the quality of correspondence matching especially in the case of smaller objects.
As an alternative, in this paper we generate an optimal descriptor space embedding given a 3D mesh model, leading to a supervised training approach.
Optimality here refers to embedding the model vertices into a descriptor space with minimal distortion of their local connectivity information.
 %while ensuring minimal distortion of local connectivity information between neighboring vertex descriptors.
 We rely on discrete exterior calculus to compute the Laplacian of the object model \cite{Crane2013a}, which generates the geometrically accurate local connectivity information. 
 We exploit this information in combination with Laplacian Eigenmaps to create the corresponding optimal descriptor space embedding.
Finally, we render target descriptor space images to input RGB images depicting the object (see Fig.~\ref{fig:illustration}). 
Ultimately, our approach generates dense object descriptors akin the original, self-supervised method without having to rely on depth information.
  
 While our approach uses more domain knowledge (3D model of the object and its known pose), it has several benefits over the original self-supervised method. 
 Primarily, we do not rely on pixel-wise correspondence matching based on noisy or lower quality consumer grade depth cameras.
 Thus, our approach can be applied straightforwardly to small, reflective, or symmetric objects, which are often found in industrial processes.
 Furthermore, we explicitly separate object and background descriptors, which avoids the problem of amodal correspondence prediction when using self-supervised training \citep{Florence2020thesis}.
 We also provide a mathematical meaning for the descriptor space, which we generate optimally in a geometrical sense irrespective of the descriptor dimension.
 We believe that, overall, this improves explainability and reliability for practitioners.

\section{Background}

This chapter briefly reviews self-supervised training of Dense Object Nets \cite{Florence2018} and nonlinear dimensionality reduction by Laplacian Eigenmaps. 

\subsection{Self-supervised Training of DONs}
\label{sec:selfsup}

To collect data for the self-supervised training approach, we use a robot and a wrist-mounted RGBD camera. 
We place the target object to an arbitrary but fixed location in the workspace of the robot. 
Then, using a quasi-random motion with the robot and the camera pointing towards the workspace, we record a scene with registered RGBD images depicting the object and its environment. 
To overcome noisy and often missing depth data of consumer grade depth sensors, all registered RGBD images are then fused into a single 3D model and depth is recomputed for each frame.
While this will improve the overall depth image quality, we noticed that in practice this also over-smooths it, which results in a loss of geometrical details particularly for small objects.
With knowledge of the object location in the fused model we can also compute the object mask in each frame.

After recording a handful of scenes with different object poses, for training we repeatedly and randomly choose two different RGB images within a scene, $I_a$ and $I_b$, to evaluate the contrastive loss. 
We use the object masks to sample $N_{m}$ corresponding and $N_{nm}$ non-corresponding pixels.
Then, the contrastive loss consists of match loss ${L}_{m}$ of corresponding pixels and non-match loss ${L}_{nm}$ of non-corresponding pixels:
\begin{align}
&{L}_{m} = \frac{1}{N_{m}}\sum_{N_{m}} D(I_a, u_a, I_b, u_b)^2, \label{eq:contrastive_m}\\
&{L}_{nm} = \frac{1}{N_{nm}}\sum_{N_{nm}} \max \left(0, M-D(I_a, u_a, I_b, u_b)\right)^2, \label{eq:contrastive_nm}\\
&{L}_{c}(I_a, I_b) = {L}_{m}(I_a, I_b) + {L}_{nm}(I_a, I_b), \label{eq:contrastive}
\end{align}
where $D(I_a, u_a, I_b, u_b) = \left\| f(I_a;\vec \theta)(u_a) - f(I_b;\vec \theta)(u_b)\right\|_2$ is the descriptor space distance between pixels $u_a$ and $u_b$ of images $I_a$ and $I_b$. 
$M\in \mathbb{R}^+$ is an arbitrarily chosen margin and $f(\cdot;\vec \theta):\mathbb{R}^{h\times w \times 3}\mapsto \mathbb{R}^{h\times w \ \times D}$ represents a fully convolutional network \cite{Long2017} with parameters $\vec \theta$.

The efficiency of the self-supervised training approach lies in the automatic generation of pixel correspondences from registered RGBD images. 
Using the object mask in image $I_a$ we can sample a pixel $u_a$ and identify corresponding pixel $u_b$ in image $I_b$ by reprojecting the depth information. 
In a similar way we can sample non-correspondences on the object and on the background. 
While this approach automatically labels tens of thousands of pixels in a single image pair, its accuracy inherently relies on the quality of the depth image.
In practice we noticed that this considerably limits the accuracy of the correspondence matching for smaller objects with consumer grade depth sensors.
For further details of the training and evaluation of DONs we refer the reader to \cite{Florence2018}.

\subsection{Laplacian Eigenmaps}
\label{sec:LE}

Assume a dataset of $N$ data points is given as $\vec X = \{\vec x_i\}_{i=1}^N$, where $\vec x\in \mathcal{M}$ lies on a manifold.
 Furthermore, we are given the connectivity information between these points as $w_{ij}\in \mathbb{R}^{\geq 0}$.
 For example, if $\vec x_i$ is a node in a graph, then we can define $w_{ij}$ to be either $1$ or $0$ given it is connected to $\vec x_j$ or not.
 The Laplacian Eigenmap method considers the problem of finding an embedding of the data points in $\vec X$ to $\{\vec y_i\}_{i=1}^N$, with $\vec y \in \mathbb{R}^D$, such that the local connectivity $w_{ij}$ measured in a Euclidean sense is preserved. 
 We can express this objective as a constrained optimization problem
\begin{eqnarray}
\vec Y^* &=& \arg \min_{\vec Y} \frac{1}{2}\sum_{j=1}^N\sum_{i=1}^Nw_{ij} 
\left\| \y_i - \y_j\right\|_2^2 \label{eq:le_obj}\\
&& \mbox{s.t.}~~\vec Y \vec C \vec Y^T = \vec I,
\end{eqnarray}
where $\vec Y = [\vec y_1, \dots, \vec y_N] \in \mathbb{R}^{D\times N}$,  $\vec C_{ii} = \sum_j \vec A_{ij}$ is a diagonal matrix with $\vec A_{ji} = \vec A_{ij} = w_{ij}$ as the connectivity matrix.
$\vec C\in \mathbb{R}^{N\times N}$ measures how strongly the data points are connected.
The constraint removes bias and enforces unit variance in each dimension of the embedding space. As it was shown in the seminal paper by \citet{Belkin2003}, this optimization problem can be expressed using the Laplacian $\vec L_{N\times N}$ as 
\begin{eqnarray}
\vec Y^* &=& \arg \min_{\vec Y} \mbox{Tr}\left( \vec Y \vec L \vec Y^T \right) \label{eq:laplaceoptim}\\ 
&&s.t.~~ \vec Y \vec C \vec Y^T = \vec I, \nonumber
\end{eqnarray}
where  $\vec L = \vec C - \vec A$ is a positive semi-definite, symmetric matrix. The solution to this optimization problem can be found by solving a generalized eigenvalue problem
\begin{equation}
\vec L \vec Y^T = \mbox{diag}(\vec \lambda) \vec C \vec Y^T,
\label{eq:geneigen}
\end{equation}
with eigenvectors corresponding to the dimensions of $\vec y$ (or rows of $\vec Y$) and non-decreasing real eigenvalues $\vec \lambda\in \mathbb{R}^{\geq 0}$. 
Interestingly, the first dimension is constant for each data point, i.e., $\vec Y_{1,:} = \mbox{const}$ and $\lambda_1 = 0$. This corresponds to embedding every data point in $\vec X$ to a single point. 
For practical reasons the very first dimension is ignored ($D=0$). 
Therefore, choosing $D=1$ corresponds to embedding $\vec X$ on a line, $D=2$ to a plane, etc. Given that the Laplacian is symmetric positive semi-definite (when we use the definition $\vec L = \vec C - \vec A$), the eigenvectors are orthogonal to each other.
Additionally, with the optimal solution we have $\mbox{diag}(\vec \lambda) = \vec Y^* \vec L \vec Y^{*T}$, that is, the eigenvalues represent the embedding error in each dimension.
Consequently, as these eigenvalues are non-decreasing, we always arrive at an optimal embedding irrespective of the dimensionality $D$.

Note that both contrastive learning and Laplacian Eigenmaps achieve a highly related objective, that is, minimizing distances in embedding space between related, or similar data.
However, they differ in how they avoid the trivial solution (mapping to a single point), either by pushing dissimilar data by a margin away, or by normalizing the solution.
Finally, the contrastive formulation gives rise to learning parametric models (e.g., neural networks), while LE directly generates the embedding from the Laplacian, which is specific to a manifold (e.g., 3D mesh).

\section{Method}
\label{sec:method}

In this Section we describe our proposed solution to generate optimal training targets for training input RGB images and describe the supervised training setup. 
We first describe how we generate an optimal descriptor space map for a 3D object that is represented by a triangle mesh and how we can handle symmetries.
Then we explain the full target image generation and supervised training pipeline.

Note that the contrastive loss of the self-supervised training approach is defined over pixel descriptors of RGB images depicting the object. 
In our case, we exploit the LE embedding directly on the mesh of the model. 
This allows us to exploit the geometry of the object and to address optimality of the embedding. 
As a second step we render the descriptor space representation of the object to generate descriptor space images for the supervised training objective.

\subsection{Embedding Object Models using Laplacian Eigenmaps}
\label{sec:eigenmaps}
We assume that a 3D model of an object is a triangle mesh consisting of $N=|V|$ vertices $V$, edges $E$ and faces $F$. We consider the mesh as a Riemannian manifold $\mathcal{M}$ embedded in $\mathbb{R}^3$. 
The vertices $\x_i \in \mathcal{M}$ are points on this manifold.
We are looking for the descriptor space maps of the vertices  $\vec y_i \in \mathbb{R}^D$.
To apply the LE solution we need to compute the Laplacian of the object model.
For triangle meshes this can be computed based on discrete exterior calculus \cite{Crane2013a}.
For an efficient and easy to implement solution we refer to \cite{Sharp2019}.
After computing the Laplacian, we can solve the same generalized eigenvalue problem as in Eq.~\eqref{eq:geneigen} to compute the $D$-dimensional descriptor space embedding of the vertices $\vec Y \in \mathbb{R}^{D\times N}$.

\begin{figure*}[th]
	\centering
	\includegraphics[width=0.175\textwidth]{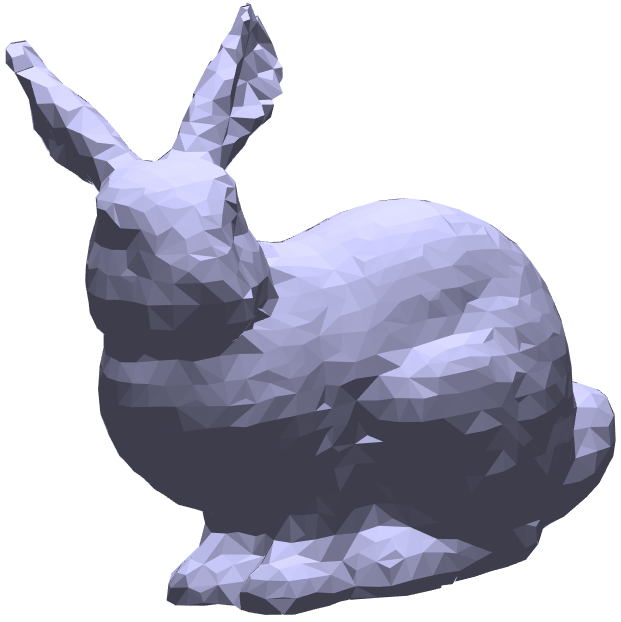}
	\includegraphics[width=0.175\textwidth]{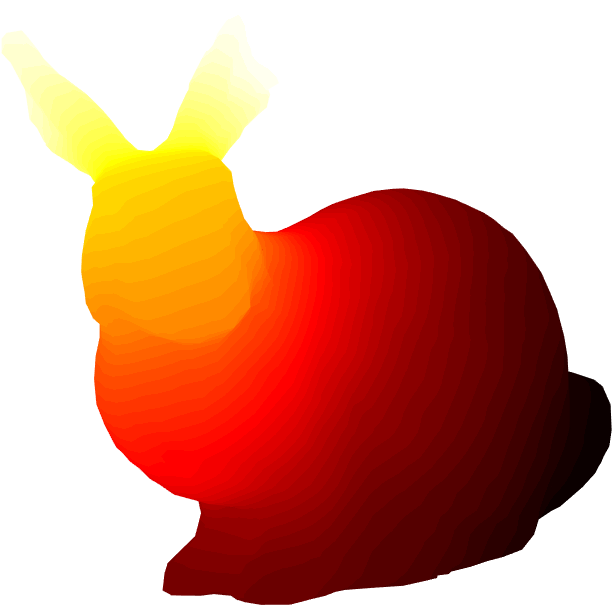}
	\includegraphics[width=0.175\textwidth]{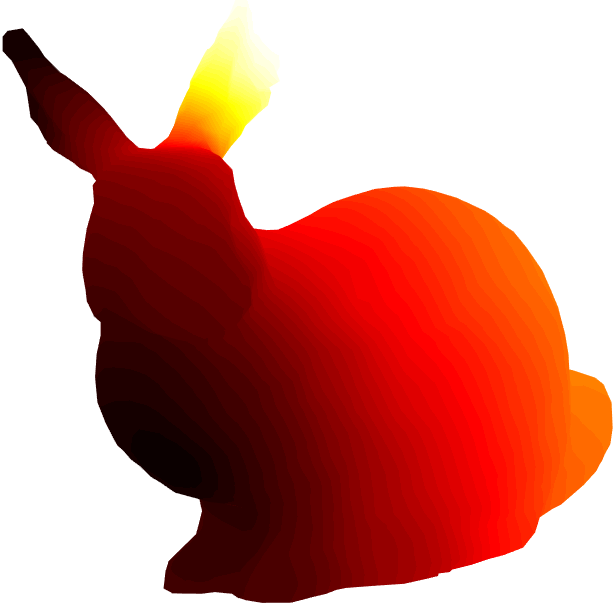}
	\includegraphics[width=0.175\textwidth]{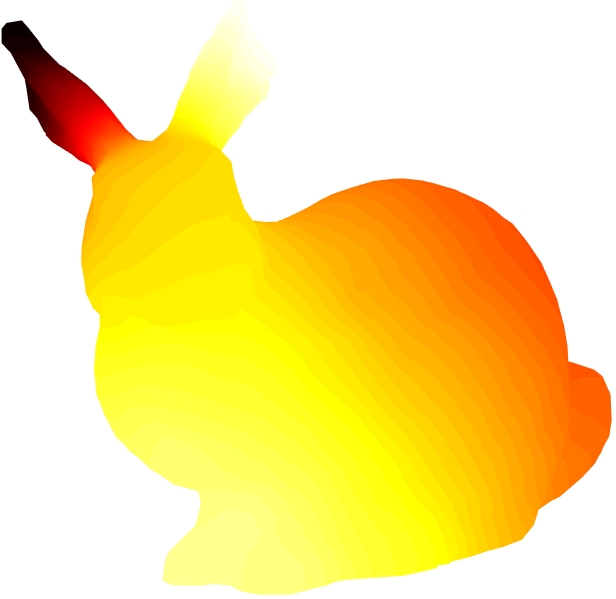}
	\includegraphics[width=0.175\textwidth]{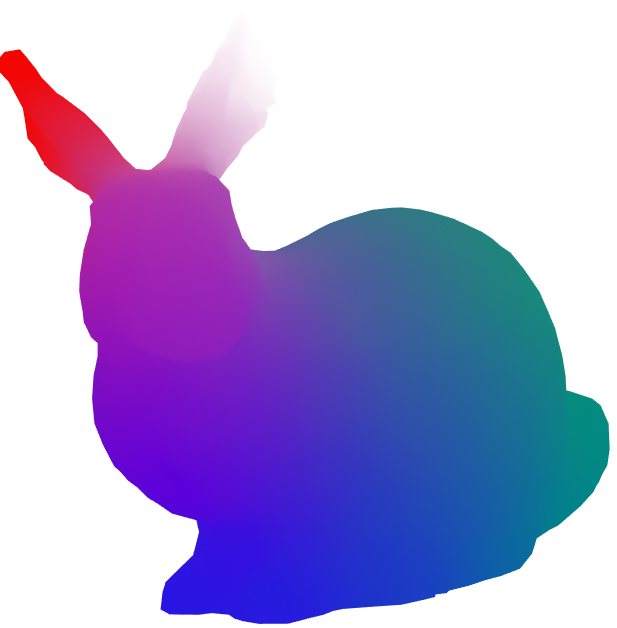}
\caption{Illustration of the Laplacian Eigenmap embedding of the Stanford bunny 
(best viewed in color). \textbf{(left):} the triangle mesh representation, 
\textbf{(middle three):} the first three eigenvector projected on the mesh, 
scaled between $[0, 1]$ and visualized with hot color map, \textbf{(right):} 
the descriptor space representation of the mesh visualized with red 
corresponding to the first eigenvector, or descriptor dimension, green for the 
second and blue for the third.}
	\vspace{-2mm}
\label{fig:bunny}
\end{figure*}

Fig.~\ref{fig:bunny} illustrates the solution of the eigenvalue problem for the 
Stanford bunny and the resulting descriptor space embedding with $D=3$. We project the solution on the vertices of the object model and use a renderer to color the faces.

\subsection{Handling Symmetry}
\label{sec:symmetry}

\begin{figure}[th]
	\centering
	\includegraphics[width=0.4\columnwidth]{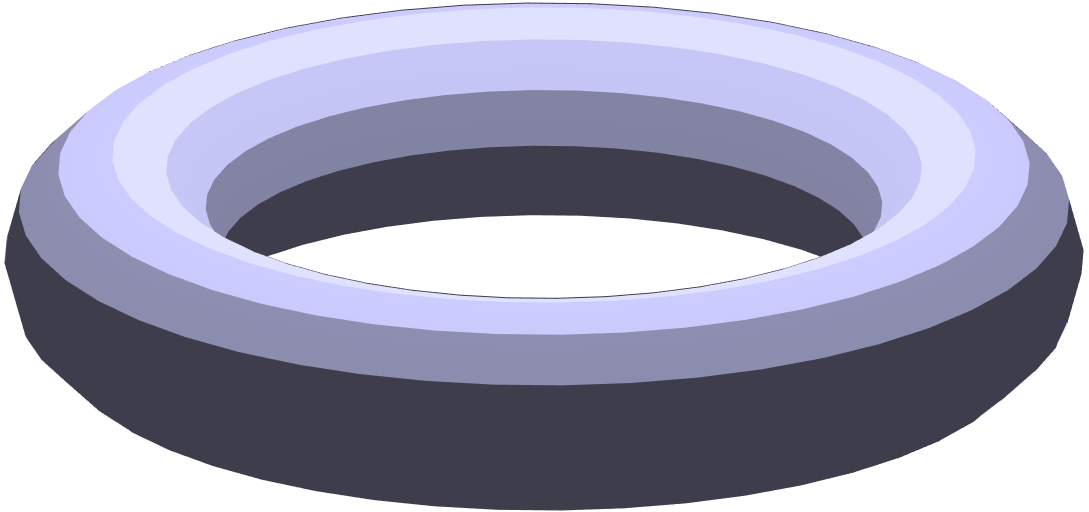}
	\hspace*{0.02\textheight}
	\includegraphics[width=0.4\columnwidth]{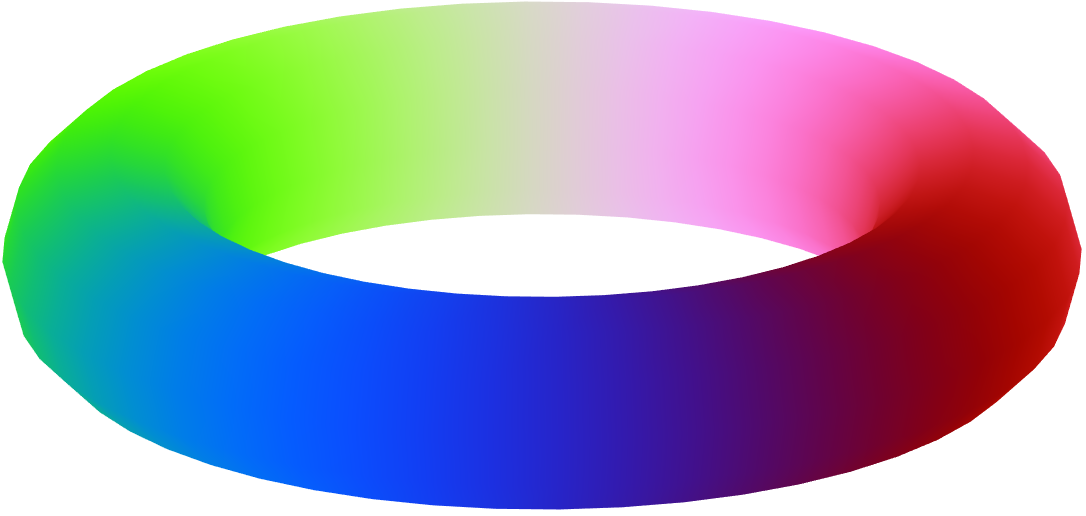}\\
	\includegraphics[width=0.4\columnwidth]{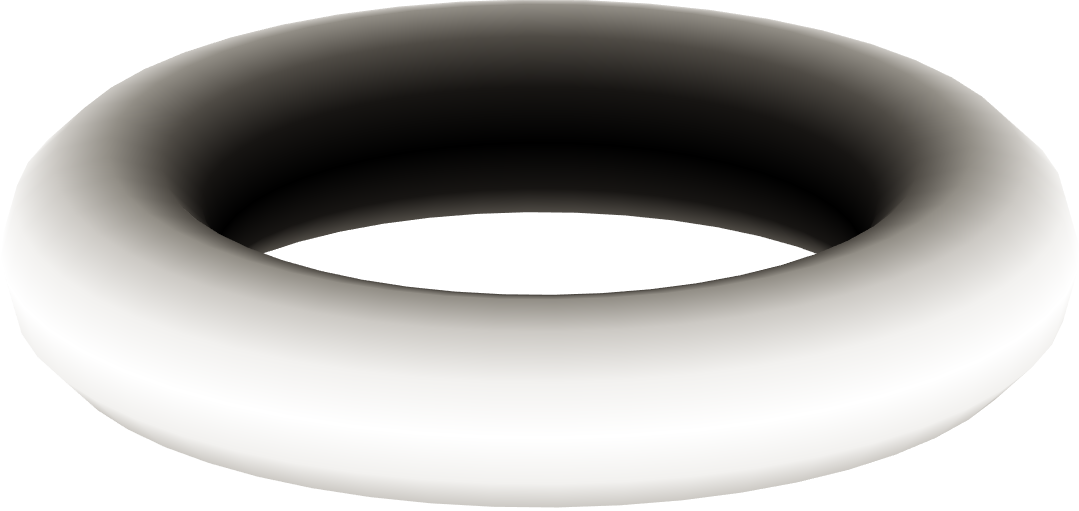}
	\caption{\textbf{(top left)} the mesh of the torus, \textbf{(top right)} 
	the asymmetric 3-dimensional descriptor projected on the mesh, 
	\textbf{(bottom)} the symmetric 3-dimensional descriptor projected on the 
	mesh.} %
	\vspace{-5mm}

	\label{fig:torus}
\end{figure}

So far we map every vertex of the mesh to a unique point in descriptor space. 
However, for objects with symmetric geometric features, typical in industrial settings, this approach will assign unique descriptors to indistinguishable vertices.
Consider the case of the torus in Fig.~\ref{fig:torus}. The mesh is invariant to rotations around the z-axis. 
If we apply the Laplacian embedding approach, we end up with descriptors that do not preserve this symmetry (top right in Fig.~\ref{fig:torus}). 
The descriptor values will be assigned purely based on the ordering of the vertices in the mesh.
Instead, we are looking for an embedding as in the bottom of Fig.~\ref{fig:torus}, which only differentiates between ``inside-outside'' vertices and which appears invariant to rotation around z-axis.

To overcome this problem, we have to (i) detect intrinsic symmetries of shapes and (ii) compress symmetric embeddings, such that symmetric vertices map to the same descriptor. 
\citet{Ovsjanikov2008} discuss an approach for detecting intrinsic symmetries for shapes represented as compact (Riemannian) manifolds. 
In particular, they showed that a shape has intrinsic symmetry if the eigenvectors of the Laplacian, that is, its Euclidean embedding appear symmetric. 
Following this result, \citet{Wang2014} defined Global Intrinsic Symmetry Invariant Functions (GISIFs) on vertices that preserve their value in the face of any homeomorphism (such as rotation around z-axis in the case of the torus). 
They also showed that such a GISIF can be composed of eigenvectors of the Laplacian. 

In particular, they propose that in case of identical consecutive eigenvalues $ \lambda_i =  \lambda_{i+1} = \dots =  \lambda_{i+L}$, such a GISIF is the squared sum of the corresponding eigenvectors $\vec y_{\mbox{sym}} = \sum_{i}^{i+L} \vec y_i^2$.   
In practice eigenvalues are rarely identical due to numerical limitations. 
To resolve this problem we heuristically consider eigenvalues in an $\epsilon$-ball as identical.

\subsection{Generating Target Images and Network Training}

Given the optimal embedding of an object model in descriptor space and the registered RGB images of a static scene, we can now generate the target descriptor space images for training.
In the collected dataset we have $K$ registered RGB images $I_i$ with camera extrinsic $\vec T^c_i \in SE(3)$, $\{I_i, \vec T^c_i\}_{i=1}^K$ depicting the object and its environment.
We compute the descriptor space embedding of the model (see Sec.~\ref{sec:eigenmaps} and \ref{sec:symmetry}) and we assume the object pose in world coordinates $\vec T^o\in SE(3)$ is known.
To generate target descriptor space images $I^d_i$ we can use a rendering engine to project the descriptor space model to the image plane of the camera, see Fig.~\ref{fig:illustration} for an illustration.
The pose of the object in the camera coordinate system can be computed from the camera poses and the object pose with $\vec T^{o,c}_i = \vec T^{c-1}_i \vec T^o$.
To generate the background image we first normalize the descriptor dimensions between $[0, 1]$. 
Then, as background we choose the descriptor which is the furthest away from object descriptors within a unit cube.
By explicitly separating object and background it becomes more unlikely to predict object descriptors in the background, which may occur when using the self-supervised training approach, also reported in \citep{Florence2020thesis}.

To train the network we rely on $\ell_2$ loss between DON output and generated descriptor space images.
However, in a given image typically the amount of pixels representing the object is significantly lower than that of the background.
Therefore, we separate the total loss into object and background loss and normalize them with the amount of pixels they occupy
\begin{align}
    \vec \theta^* =& \arg \min_{\vec \theta} \sum_{i=1}^K \Big( \frac{1}{P_{i,obj}}\left\| M_{i,obj} \circ \left( f(I_i;\vec \theta) - I^d_i \right)\right\|^2_2 + \nonumber \\
                   & \frac{1}{P_{i,back}}\left\| M_{i,back} \circ \left( f(I_i;\vec \theta) - I^d_i \right) \right\|_2^2 \Big),
\label{eq:l2loss}
\end{align}
where $\vec \theta$ are the network parameters, $P_{i,obj/back}$ is the amount of pixels the object, or background occupies in the image $I_i$ and $M_{i,obj/back}$ is the object or background mask. 
For convenience here we overloaded the notation $M$, which refers to margin in the self-supervised training approach (Sec.~\ref{sec:selfsup}) and mask here. 
The mask can be generated via rendering the object in the image plane, similar to how we generate target images $I^d_i$.

\begin{figure}
	\centering
	\includegraphics[width=\linewidth]{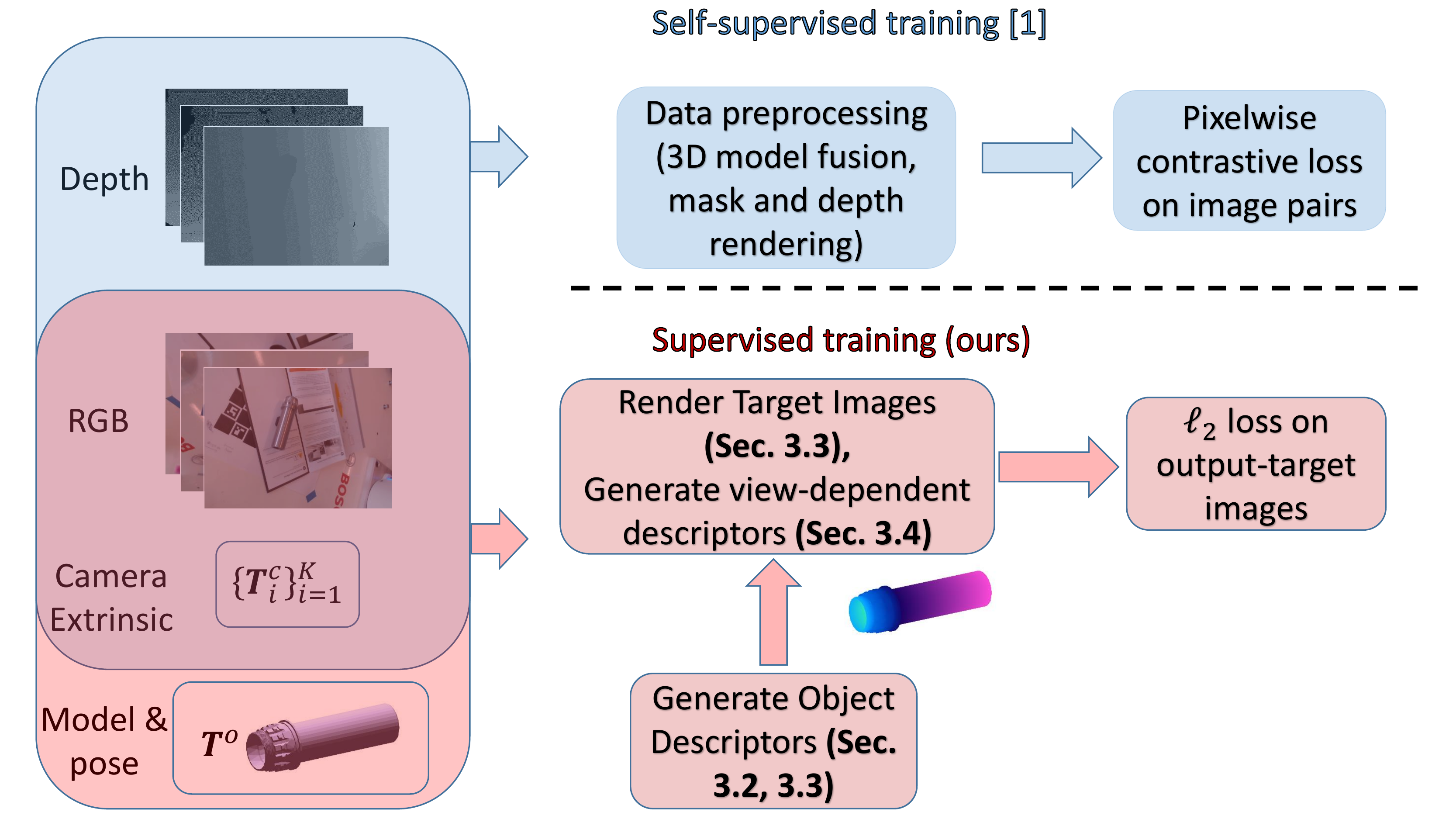}
    \caption{Comparison of the self-supervised and the supervised training (ours) pipelines. 
        The first utilizes depth images to find pixel correspondences and relies on the contrastive pixel-wise loss. 
		Ours exploits the object model and its pose in world coordinates to render training target images, which will be used in the loss function (Eq.~\eqref{eq:l2loss}).}
	\label{fig:pipelines}
\end{figure}

We compare the training pipeline of the original self-supervised DON framework \cite{Florence2018} and our proposed supervised approach in Fig.~\ref{fig:pipelines}.
We use the same static scene recording with a wrist-mounted camera for both training approaches. 
While the self-supervised training approach takes the depth, RGB, and camera extrinsic parameters, ours takes the model pose in world coordinates and its mesh instead of depth images.
Then, both approaches perform data preprocessing, 3D model fusion, depth and mask rendering for the contrastive loss, object descriptor generation and descriptor image rendering for the $\ell_2$ loss.
Finally, both training approaches optimize the network parameters.

\subsection{View-dependent Descriptors}

Note that our framework can be extended with additional descriptor dimensions that not necessarily minimize the embedding objective defined in Eq.~\eqref{eq:le_obj}.
In practice we noticed that the self-supervised training approach learns descriptor  dimensions which mask the object and its edge resulting in \emph{view-dependent} descriptors.
This is likely a consequence of the self-supervised training adapting to slightly inaccurate pixel correspondence generation in the face of camera extrinsic errors and oversmoothed depth images.
As our training method relies on the same registered RGBD scene recording, to 
improve training robustness we considered two view-dependent descriptors as an 
extension to our optimally generated geometric descriptors. 

\begin{figure}
	\centering
	\includegraphics[width=\linewidth]{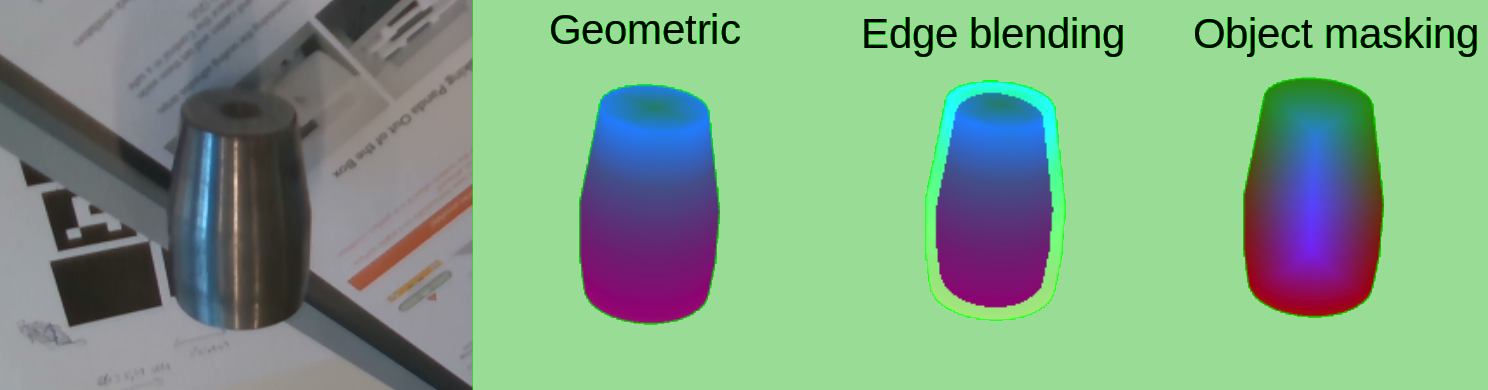}
	\caption{Visualization of view-dependent descriptors. On the left side we can see the original RGB image, depicting the toolcap. Using $D=3$ we generate the optimal geometric descriptor image. Then, we can either blend the object edge with the background, or generate an object mask.}
	\label{fig:viewdependent}
	\vspace{-2mm}	
\end{figure}

First, we considered blending descriptors on the edge of the object with the background descriptor. 
This mimics a view-dependent edge detection descriptor that separates object and background.
Second, we can use a descriptor dimension as object mask that is $0$ outside the object and gradually increases to $1$ at the center of the object in the image plane.
The former option smooths the optimal descriptor images around the edges, while the latter extends or replaces descriptor dimensions with an object mask.
These two additional descriptors together with the optimal geometric descriptors are shown in Fig.~\ref{fig:viewdependent} for a metallic toolcap object.
Note that these features can be computed automatically from the object mask of the current frame.

We experimented with these extensions to improve grasp pose prediction tasks in practice. 
This extension shows on one hand the flexibility of using our approach by manipulating descriptor dimensions.
On the other hand, it highlights the benefits of using the self-supervised training approach to automatically address inaccuracies in the data.

\section{Evaluation}
\label{sec:evaluation}
In this section we provide a comparison of the trained networks. 
While the DON framework can be applied for different robotic applications, in this paper we consider the oriented grasping of industrial work pieces. 
To this end we first provide a quantitative evaluation of the prediction accuracy of the networks. 
We measure accuracy by how consistently the network predicts descriptors for specific points on the object from different points of view.
Then, we derive a grasp pose prediction pipeline and use it in a real-world robot grasp and placement scenario.

\subsection{Hardware, Network and Scene Setup}
\begin{figure}[h]
	\centering
	\includegraphics[width=0.85\columnwidth]{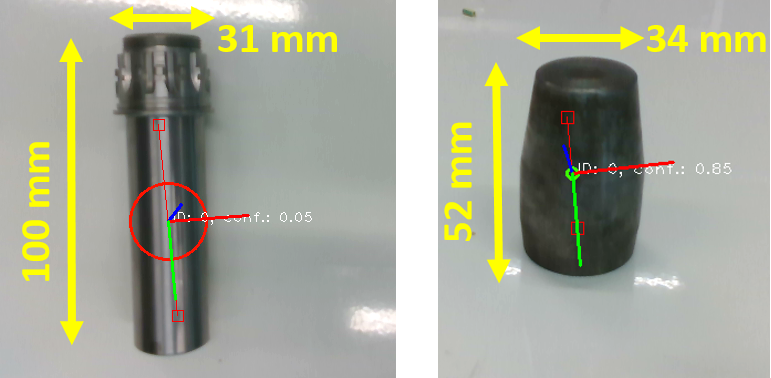}
	\caption{The metallic shaft and toolcap objects with predicted grasp poses by tracking $2$ descriptors, or points on the objects.}
	\label{fig:objects}
\end{figure}

We use a Franka Emika Panda 7-DoF robot with a parallel gripper \cite{panda}, equipped with a wrist-mounted Intel RealSense D435 camera. 
The relative transformation from end-effector to camera coordinate system is computed with ChArUco board calibration \cite{charuco} and forward kinematics of the robot. 
We record RGB and depth images of size $640 \times 480$ and at $100 \mu m$ resolution at 15Hz.
For a given scene we record $\sim 500$ images with a precomputed robot end-effector trajectory resulting in highly varying camera poses.
Our target objects are a metallic shaft and a toolcap, both of which appear invariant to rotation around one axis (see Fig.~\ref{fig:objects}).

For both, self-supervised and supervised training, we use descriptor dimension $D=3$, consider only a single object per scene, and train with background randomization.
For the supervised approach we consider two ways of generating data. 
We first compute the optimal embedding, then we either use edge blending, or object masking. 
In the latter case we use the first two geometric descriptor dimensions and add the view-dependent object mask as the third (see Fig.~\ref{fig:viewdependent}). 
For the self-supervised approach we use the hyper-parameters defined in \cite{Florence2018}. 
That is, we use $M=0.5$ as margin for on-object non-match loss, $M=2.5$ for background non-match loss. 
We do not use normalization of the descriptors and use scaling by hard-negatives.
The network structure for both method is ResNet-34 pretrained on ImageNet.

\subsection{Quantitative Evaluation of Prediction Accuracy}

\begin{figure}[h]
	\centering
	\includegraphics[width=0.99\columnwidth]{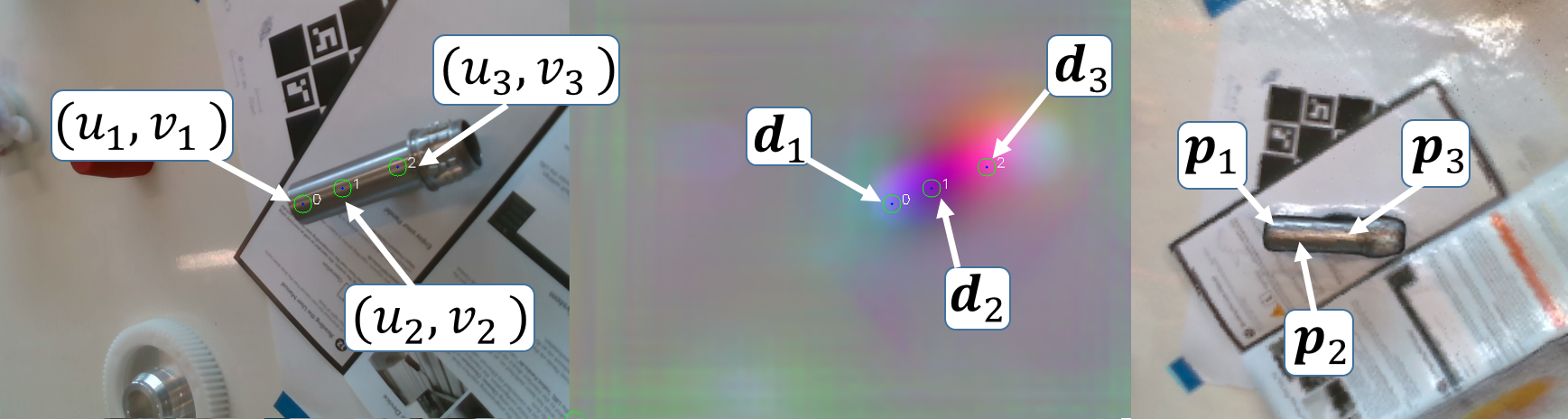}
	\caption{Given a scene during evaluation, we select a few pixels on the RGB image corresponding to points on an object (shown for $3$ points). 
	We record their descriptors and world coordinates as a tracking reference. }
	%Then, we iterate over every image in  the scene to evaluate tracking accuracy.}
	\label{fig:evaluation}
\end{figure}

We now provide an evaluation for how accurately the networks can track points on an object from different views. 
For a given static scene we choose an RGB image $I$ and select an arbitrary amount of reference pixels \emph{on the object} $(u_i, v_i),~i=1,\dots,N$ (see  Fig.~\ref{fig:evaluation} left).
Then, we record the 3D world coordinates of the pixels $\vec p_i\in\mathbb{R}^3$ using the registered depth images and their corresponding  descriptors $\vec d_i = I^d(u_i, v_i) \in \mathbb{R}^D$, where $I^d = f(I;\vec \theta)$ is the descriptor image evaluated by the network (see Fig.~\ref{fig:evaluation} middle and right).
Then, we iterate over every image $I_j,~j=1,\dots,M$ in the scene and evaluate the network $I^d_j = f(I_j;\vec \theta)$.
In every descriptor space image $I^d_j$ we find the closest descriptors to our reference set $\vec d_i$ and their pixel location $(u_i, v_i)$ by solving $(u_i, v_i)^*_j = \arg \min_{u_i, v_i} \left\| I^d_j(u_i, v_i) - \vec d_i \right\|_2^2,~\forall j$. 
We only consider those reference points which are visible and have valid depth data.
Using the registered depth images and pixel values $(u_i, v_i)^*_j$ we can compute the predicted world coordinates of the points $\tilde{\vec p}_i$ and the tracking error $\left\|\tilde{\vec p}_i - \vec p_i\right\|_2$ for each frame $j$ and for each chosen descriptor $i$. 

\begin{figure}[h]
	\centering
	\includegraphics[width=0.99\columnwidth]{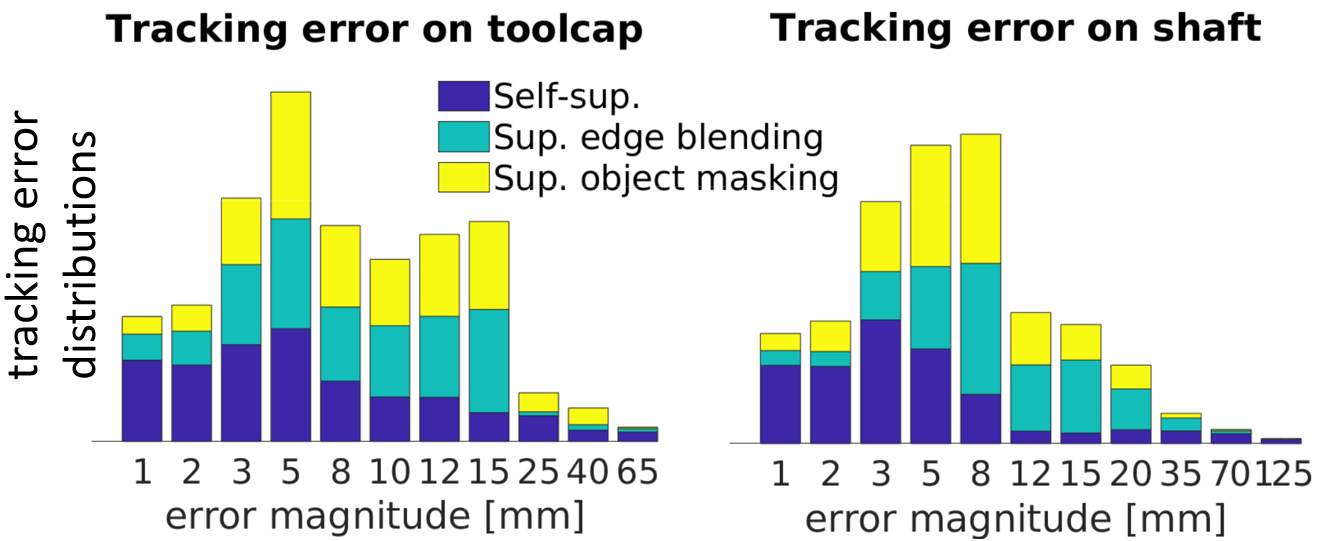}
	\caption{Tracking error distributions evaluated on every scene for both objects. 
        Note the log-scale of the $x$-axis. 
        While self-supervised training provides higher accuracy, it has a long-tail error distribution, which sometimes leads to significantly higher errors. 
        The supervised approaches have a slightly lower accuracy, but increased robustness.}
	\label{fig:error_toolcap_shaft}
\end{figure}

Fig.~\ref{fig:error_toolcap_shaft} shows the tracking error distributions of the supervised and self-supervised approaches. 
For improved visualization, the normalized distributions are summarized with stacked histograms and with log-scale on the $x$ axis.
In general, the self-supervised training method seems more accurate, but we notice a long-tail error distribution that can lead to significant prediction errors. 
The supervised approaches provide better robustness without the long-tail distribution, albeit being slightly less accurate.
We noticed that the self-supervised training leads to a descriptor space that often mistakes parts of the background as the object, which could lead to higher tracking errors. 
With our supervised training method this occurs less frequently as we explicitly separate object and background in descriptor space, while this separation is less apparent in the self-supervised case.

\subsection{Grasp Pose Prediction}
\label{sec:grasp_point_prediction}

The dense descriptor representation gives rise to a variety of grasp pose configurations. 
In the simplest case, consider a 3D top-down grasp: choose a descriptor $\vec d$ that corresponds to a specific point on the object.
Then, in a new scene with a registered RGBD image, evaluate the network and find the closest descriptor $\vec d$ with corresponding pixel values $(u, v)$.
Finally, project the depth value at $(u, v)$ in world coordinates to identify the grasp point $\tilde{\vec p} \in \mathbb{R}^3$. 
This method previously described in \citet{Florence2018} together with grasp pose optimization.

Following this method, we can also encode orientation by defining an axis grasp. 
For this purpose we identify $2$ descriptors $\vec d_1, \vec d_2$ and during prediction find the corresponding world coordinates $\tilde{\vec p}_1, \tilde{\vec p}_2$.
To define the grasp position we use a linear combination of $\tilde{\vec p}_1, \tilde{\vec p}_2$, e.g. taking their mean.
To define the grasp orientation we align a given axis with the direction $\tilde{\vec p}_1 - \tilde{\vec p}_2$ and choose an additional axis arbitrarily, e.g. align it with the z-axis of the camera, or world-coordinates. 
The third axis can then be computed by taking the cross product. 
See Fig.~\ref{fig:objects} for predicted axis-grasps on two test objects.
Finally, to fully specify a given 6D pose we can extend the above technique to track $3$ or more descriptors.

For our test objects we have chosen an axis grasp representation by tracking $2$ descriptors. 
We align the x-axis of the grasp pose with the predicted points and we choose as z-axis the axis orthogonal to the image plane of the camera.

\begin{figure}[h]
	\centering
	\includegraphics[width=0.99\columnwidth]{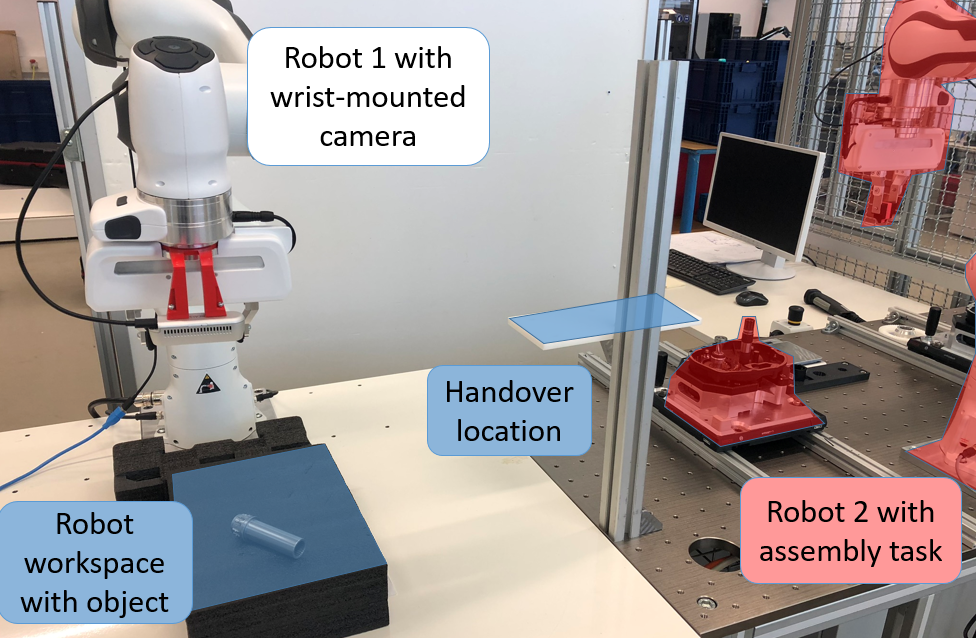}
	\caption{The summary of the experiment setup.}
	\label{fig:setup}
\end{figure}

\textbf{Experiment setup.} 
Consider an industrial setting where a human places an object with an arbitrary position and orientation in the workspace of a robot.
The robot's goal is to grasp the object and place it at a handover location, such that another, pre-programmed robot (without visual feedback) can grasp and further manipulate the object in an assembly task.
The setup is shown in Fig.~\ref{fig:setup}.

For real robot grasping evaluations we have chosen the toolcap as our test object.
We observed that the experiment with shaft grasping was not challenging for any of the networks.
However, we noticed that due to the smaller size of the toolcap and its symmetric features the self-supervised training approach was pushed to the limit in terms of accuracy.
To reduce variance, we have chosen $8$ \emph{fixed} poses for the toolcap and repeated the experiment for each trained DON on each of the $8$ poses. 
We categorize the experiment outcome as follows:
\begin{itemize}
\item \texttt{fail:} if the grasp was unsuccessful, the toolcap remains untouched
\item \texttt{grasp only:} if the grasp was successful, but placement failed (e.g., the toolcap was dropped, or was grasped with the wrong orientation)
\item \texttt{grasp \& place:} if grasping and placement with the right orientation were successful
\end{itemize}

\begin{figure}[h]
	\centering
	\includegraphics[width=0.65\columnwidth]{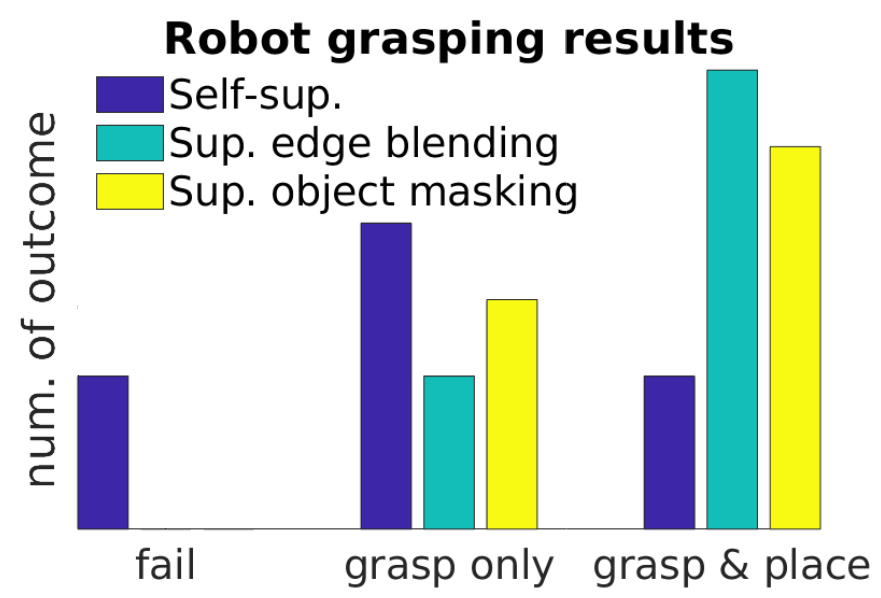}
	\caption{Summary of the real robot experiments. 
	 }
	\label{fig:robotgrasp}
\end{figure}

Fig.~\ref{fig:robotgrasp} illustrates results of the experiment. 
The self-supervised training resulted in grasps that are often misaligned, which could lead to a failed placement or even a failed grasp. 
The supervised training approaches show better robustness with at least a successful grasp in all the cases.
Most failed placements were due flipped orientations leading to upside-down final poses.
Overall, the results are highly consistent with our previous evaluation on reference point tracking accuracy.
The supervised trained approaches provided more robust performance.

\section{Related Work}

 \label{sec:related_work}

Dense descriptor learning via DONs was popularized by \citet{Florence2018} for the robotics community, offering a flexible perceptual world representation with self-supervised training, which is also applicable to non-rigid objects. 
Recently, this work was extended to learn descriptors in dynamic scenes for visuo-motor policy learning \cite{Florence2020}. 
\citet{Manuelli2020} propose to define model predictive controllers based on tracking key descriptors, similar to our grasp pose prediction pipeline.
\citet{Sundaresan2020} showed how ropes can be manipulated based on dense descriptor learning in simulation.
Similar to these approaches \citet{Vecerik2020} propose to learn and track human annotated keypoints for robotic cable plugging.
Visual representation by dense descriptor is not novel though. 
\citet{Choy2016} introduced a deep learning framework for dense correspondence learning of image features using fully convolutional neural networks by \citet{Long2017}.
Later, \citet{Schmidt2017} showed how self-supervised training can be incorporated when learning descriptors from video streams.

 Recently, several authors used dense object descriptors for 6D pose estimation. 
 \citet{Zakharov2019} generate 2D or 3D surface features (colors) of objects via simple spherical or cylindrical projections. 
 Then, they train a network in a supervised manner to predict these dense feature descriptors and a mask of the object. 
 For 6D pose estimation, they project the colored mesh into the predicted feature image. 
 \citet{Periyasamy2019} and \citet{Li2018} propose similar training procedures to estimate dense feature descriptors. 
 \citet{Periyasamy2019} do not report details for generating the dense descriptors and use differentiable rendering for object pose estimation. 
 \citet{Li2018} use relative translations of the object as dense feature descriptors and a combination of rendering and an additional refinement network for pose estimation. 
  The supervised training procedure proposed in all of the aforementioned methods is similar to ours. 
  The main difference is our novel method of computing optimal dense descriptors that locally preserves geometric properties. 
 
 Our work also borrows ideas from the geometry processing community.
 \citet{Lee2005} consider embedding data manifolds with non-linear dimensionality reduction that have specific geometric features.
 \citet{Liu2012} propose a spectral processing framework that also works with point clouds to compute a discrete Laplacian.
The works by \citet{Crane2013a, Crane2017} build on discrete exterior calculus to efficiently compute the LB operator for meshes, which we also use in this work.
Finally, detecting and processing symmetrical features of 3D objects represented as Riemannian manifolds is considered by the works of \citet{Ovsjanikov2008, Wang2014}.

 \section{Summary}
In this paper we presented a supervised training approach for Dense Object Nets from registered RGB camera streams without depth information. 
We showed that by embedding an object model into an optimally generated descriptor space we achieve a similar objective as in the self-supervised case.
Our experiments show increased grasp pose prediction robustness for smaller, metallic objects. 
While self-supervised training has obvious benefits, our supervised method improved results for consumer grade cameras by incorporating domain knowledge suitable for industrial processes.

Future work will investigate how to extend supervised training with multiple object classes efficiently and a wider range of applications. For example,  6D pose estimation from RGB images, region of interest detection and generalizing grasp pose prediction to multiple object instances and classes. 

\clearpage

%===============================================================================

\bibliography{main_preprint}  % .bib

\begin{thebibliography}{25}
\providecommand{\natexlab}[1]{#1}
\providecommand{\url}[1]{\texttt{#1}}
\providecommand{\urlprefix}{URL }
\expandafter\ifx\csname urlstyle\endcsname\relax
  \providecommand{\doi}[1]{doi:\discretionary{}{}{}#1}\else
  \providecommand{\doi}{doi:\discretionary{}{}{}\begingroup
  \urlstyle{rm}\Url}\fi

\bibitem[{Belkin and Niyogi(2003)}]{Belkin2003}
Belkin, M.; and Niyogi, P. 2003.
\newblock {Laplacian eigenmaps for dimensionality reduction and data
  representation}.
\newblock \emph{Neural Computation} 15(6): 1373--1396.
\newblock ISSN 08997667.
\newblock \doi{10.1162/089976603321780317}.

\bibitem[{Choy et~al.(2016)Choy, Gwak, Savarese, and Chandraker}]{Choy2016}
Choy, C.~B.; Gwak, J.; Savarese, S.; and Chandraker, M. 2016.
\newblock {Universal Correspondence Network}.
\newblock In \emph{Advances in Neural Information Processing Systems 30}.

\bibitem[{Crane et~al.(2013)Crane, de~Goes, Desbrun, and
  Schr\"{o}der}]{Crane2013a}
Crane, K.; de~Goes, F.; Desbrun, M.; and Schr\"{o}der, P. 2013.
\newblock Digital Geometry Processing with Discrete Exterior Calculus.
\newblock In \emph{ACM SIGGRAPH 2013 courses}, SIGGRAPH '13. New York, NY, USA:
  ACM.

\bibitem[{Crane, Weischedel, and Wardetzky(2017)}]{Crane2017}
Crane, K.; Weischedel, C.; and Wardetzky, M. 2017.
\newblock {The heat method for distance computation}.
\newblock \emph{Communications of the ACM} 60(11): 90--99.
\newblock ISSN 15577317.
\newblock \doi{10.1145/3131280}.

\bibitem[{Florence, Manuelli, and Tedrake(2018)}]{Florence2018}
Florence, P.; Manuelli, L.; and Tedrake, R. 2018.
\newblock {Dense Object Nets: Learning Dense Visual Object Descriptors By and
  For Robotic Manipulation}.
\newblock \emph{Conference on Robot Learning} .

\bibitem[{Florence, Manuelli, and Tedrake(2020)}]{Florence2020}
Florence, P.; Manuelli, L.; and Tedrake, R. 2020.
\newblock {Self-Supervised Correspondence in Visuomotor Policy Learning}.
\newblock \emph{IEEE Robotics and Automation Letters} 5(2): 492--499.
\newblock ISSN 23773766.
\newblock \doi{10.1109/LRA.2019.2956365}.

\bibitem[{Florence(2020)}]{Florence2020thesis}
Florence, P.~R. 2020.
\newblock \emph{Dense Visual Learning for Robot Manipulation}.
\newblock Ph.D. thesis, Massachusetts Institute of Technology.

\bibitem[{Franka(2018)}]{panda}
Franka, E. 2018.
\newblock Panda Arm.
\newblock \url{https://www.franka.de/panda/}.

\bibitem[{Garrido-Jurado et~al.(2014)Garrido-Jurado, Mu\~{n}oz Salinas,
  Madrid-Cuevas, and Mar\'{\i}n-Jim\'{e}nez}]{charuco}
Garrido-Jurado, S.; Mu\~{n}oz Salinas, R.; Madrid-Cuevas, F.; and
  Mar\'{\i}n-Jim\'{e}nez, M. 2014.
\newblock Automatic Generation and Detection of Highly Reliable Fiducial
  Markers under Occlusion.
\newblock \emph{Pattern Recogn.} 47(6): 2280–2292.
\newblock ISSN 0031-3203.
\newblock \doi{10.1016/j.patcog.2014.01.005}.
\newblock \urlprefix\url{https://doi.org/10.1016/j.patcog.2014.01.005}.

\bibitem[{Hadsell, Chopra, and LeCun(2006)}]{Hadsell2006}
Hadsell, R.; Chopra, S.; and LeCun, Y. 2006.
\newblock {Dimensionality reduction by learning an invariant mapping}.
\newblock \emph{Proceedings of the IEEE Computer Society Conference on Computer
  Vision and Pattern Recognition} 2: 1735--1742.
\newblock ISSN 10636919.
\newblock \doi{10.1109/CVPR.2006.100}.

\bibitem[{Lee and Verleysen(2005)}]{Lee2005}
Lee, J.~A.; and Verleysen, M. 2005.
\newblock {Nonlinear dimensionality reduction of data manifolds with essential
  loops}.
\newblock \emph{Neurocomputing} 67(1-4 SUPPL.): 29--53.
\newblock ISSN 09252312.
\newblock \doi{10.1016/j.neucom.2004.11.042}.

\bibitem[{Li et~al.(2018)Li, Wang, Ji, Xiang, and Fox}]{Li2018}
Li, Y.; Wang, G.; Ji, X.; Xiang, Y.; and Fox, D. 2018.
\newblock Deepim: Deep iterative matching for 6d pose estimation.
\newblock In \emph{Proceedings of the European Conference on Computer Vision
  (ECCV)}, 683--698.

\bibitem[{Liu, Prabhakaran, and Guo(2012)}]{Liu2012}
Liu, Y.; Prabhakaran, B.; and Guo, X. 2012.
\newblock {Point-based manifold harmonics}.
\newblock \emph{IEEE Transactions on Visualization and Computer Graphics}
  18(10): 1693--1703.
\newblock ISSN 10772626.
\newblock \doi{10.1109/TVCG.2011.152}.

\bibitem[{Manuelli et~al.(2020)Manuelli, Li, Florence, and
  Tedrake}]{Manuelli2020}
Manuelli, L.; Li, Y.; Florence, P.; and Tedrake, R. 2020.
\newblock {Keypoints into the Future: Self-Supervised Correspondence in
  Model-Based Reinforcement Learning}.
\newblock \emph{Conference on Robot Learning}
  \urlprefix\url{http://arxiv.org/abs/2009.05085}.

\bibitem[{Ovsjanikov, Sun, and Guibas(2008)}]{Ovsjanikov2008}
Ovsjanikov, M.; Sun, J.; and Guibas, L. 2008.
\newblock {Global intrinsic symmetries of shapes}.
\newblock \emph{Computer Graphics Forum} 27(5): 1341--1348.
\newblock ISSN 14678659.
\newblock \doi{10.1111/j.1467-8659.2008.01273.x}.

\bibitem[{Periyasamy, Schwarz, and Behnke(2019)}]{Periyasamy2019}
Periyasamy, A.~S.; Schwarz, M.; and Behnke, S. 2019.
\newblock Refining 6D Object Pose Predictions using Abstract
  Render-and-Compare.
\newblock \emph{arXiv preprint arXiv:1910.03412} .

\bibitem[{Schmidt, Newcombe, and Fox(2017)}]{Schmidt2017}
Schmidt, T.; Newcombe, R.; and Fox, D. 2017.
\newblock {Self-Supervised Visual Descriptor Learning for Dense
  Correspondence}.
\newblock \emph{IEEE Robotics and Automation Letters} 2(2): 420--427.
\newblock ISSN 2377-3766.
\newblock \doi{10.1109/LRA.2016.2634089}.

\bibitem[{Sharp, Soliman, and Crane(2019)}]{Sharp2019}
Sharp, N.; Soliman, Y.; and Crane, K. 2019.
\newblock {The Vector Heat Method}.
\newblock \emph{ACM Transactions on Graphics} 38(3): 1--19.
\newblock ISSN 07300301.
\newblock \doi{10.1145/3243651}.

\bibitem[{Shelhamer, Long, and Darrell(2017)}]{Long2017}
Shelhamer, E.; Long, J.; and Darrell, T. 2017.
\newblock Fully Convolutional Networks for Semantic Segmentation.
\newblock \emph{IEEE Trans. Pattern Anal. Mach. Intell.} 39(4): 640–651.
\newblock ISSN 0162-8828.
\newblock \doi{10.1109/TPAMI.2016.2572683}.
\newblock \urlprefix\url{https://doi.org/10.1109/TPAMI.2016.2572683}.

\bibitem[{Sundaresan et~al.(2020)Sundaresan, Grannen, Thananjeyan, Balakrishna,
  Laskey, Stone, Gonzalez, and Goldberg}]{Sundaresan2020}
Sundaresan, P.; Grannen, J.; Thananjeyan, B.; Balakrishna, A.; Laskey, M.;
  Stone, K.; Gonzalez, J.~E.; and Goldberg, K. 2020.
\newblock {Learning Rope Manipulation Policies Using Dense Object Descriptors
  Trained on Synthetic Depth Data} (3).
\newblock \urlprefix\url{http://arxiv.org/abs/2003.01835}.

\bibitem[{Tenenbaum, {De Silva}, and Langford(2000)}]{Tenenbaum2000}
Tenenbaum, J.~B.; {De Silva}, V.; and Langford, J.~C. 2000.
\newblock {A global geometric framework for nonlinear dimensionality
  reduction}.
\newblock \emph{Science} 290(5500): 2319--2323.
\newblock ISSN 00368075.
\newblock \doi{10.1126/science.290.5500.2319}.

\bibitem[{{Van Der Maaten}, Postma, and {Van Den
  Herik}(2009)}]{VanDerMaaten2009}
{Van Der Maaten}, L. J.~P.; Postma, E.~O.; and {Van Den Herik}, H.~J. 2009.
\newblock {Dimensionality Reduction: A Comparative Review}.
\newblock \emph{Journal of Machine Learning Research} 10: 1--41.
\newblock ISSN 0169328X.
\newblock \doi{10.1080/13506280444000102}.

\bibitem[{Vecerik et~al.(2020)Vecerik, Regli, Sushkov, Barker, Pevceviciute,
  Roth{\"{o}}rl, Schuster, Hadsell, Agapito, and Scholz}]{Vecerik2020}
Vecerik, M.; Regli, J.-B.; Sushkov, O.; Barker, D.; Pevceviciute, R.;
  Roth{\"{o}}rl, T.; Schuster, C.; Hadsell, R.; Agapito, L.; and Scholz, J.
  2020.
\newblock {S3K: Self-Supervised Semantic Keypoints for Robotic Manipulation via
  Multi-View Consistency}.
\newblock \emph{Conference on Robot Learning}
  \urlprefix\url{http://arxiv.org/abs/2009.14711}.

\bibitem[{Wang et~al.(2014)Wang, Simari, Su, and Zhang}]{Wang2014}
Wang, H.; Simari, P.; Su, Z.; and Zhang, H. 2014.
\newblock {Spectral global intrinsic symmetry invariant functions}.
\newblock \emph{Proceedings - Graphics Interface} 209--215.
\newblock ISSN 07135424.

\bibitem[{Zakharov, Shugurov, and Ilic(2019)}]{Zakharov2019}
Zakharov, S.; Shugurov, I.; and Ilic, S. 2019.
\newblock Dpod: 6d pose object detector and refiner.
\newblock In \emph{Proceedings of the IEEE International Conference on Computer
  Vision}, 1941--1950.

\end{thebibliography}

\end{document}